\documentclass[10pt,twocolumn,letterpaper]{article}

\usepackage{iccv}
\usepackage{times}
\usepackage{epsfig}
\usepackage{graphicx}
\usepackage{amsmath}
\usepackage{amssymb}
\usepackage{cite}

\usepackage[pagebackref=false,breaklinks=true,letterpaper=true,colorlinks,bookmarks=false]{hyperref}

\DeclareMathOperator{\iou}{IoU}

\iccvfinalcopy 

\def\iccvPaperID{1369} 

\ificcvfinal  
\begin{document}
\graphicspath{{figs/}}

\title{Incremental Learning of Object Detectors without Catastrophic Forgetting}

\author{\vspace{-0.5cm}Konstantin Shmelkov
\and Cordelia Schmid\vspace{0.2cm}\\
\large{Inria\thanks{Univ.\ Grenoble Alpes, Inria, CNRS, Grenoble INP, LJK, 38000 Grenoble, France.}\vspace{-0.2cm}}
\and Karteek Alahari
}

\maketitle

\begin{abstract}
\vspace{-0.3cm}
Despite their success for object detection, convolutional neural networks are
ill-equipped for incremental learning, i.e., adapting the original model
trained on a set of classes to additionally detect objects of new classes, in
the absence of the initial training data. They suffer from ``catastrophic
forgetting''---an abrupt degradation of performance on the original set of
classes, when the training objective is adapted to the new classes. We present
a method to address this issue, and learn object detectors incrementally, when
neither the original training data nor annotations for the original classes in
the new training set are available. The core of our proposed solution is a loss
function to balance the interplay between predictions on the new classes and a
new distillation loss which minimizes the discrepancy between responses for old
classes from the original and the updated networks. This incremental learning
can be performed multiple times, for a new set of classes in each step, with a
moderate drop in performance compared to the baseline network trained on the
ensemble of data. We present object detection results on the PASCAL VOC 2007
and COCO datasets, along with a detailed empirical analysis of the approach.
\vspace{-0.2cm}
\end{abstract}

\begin{figure}[t]
\begin{center}
\includegraphics[width=0.8\linewidth]{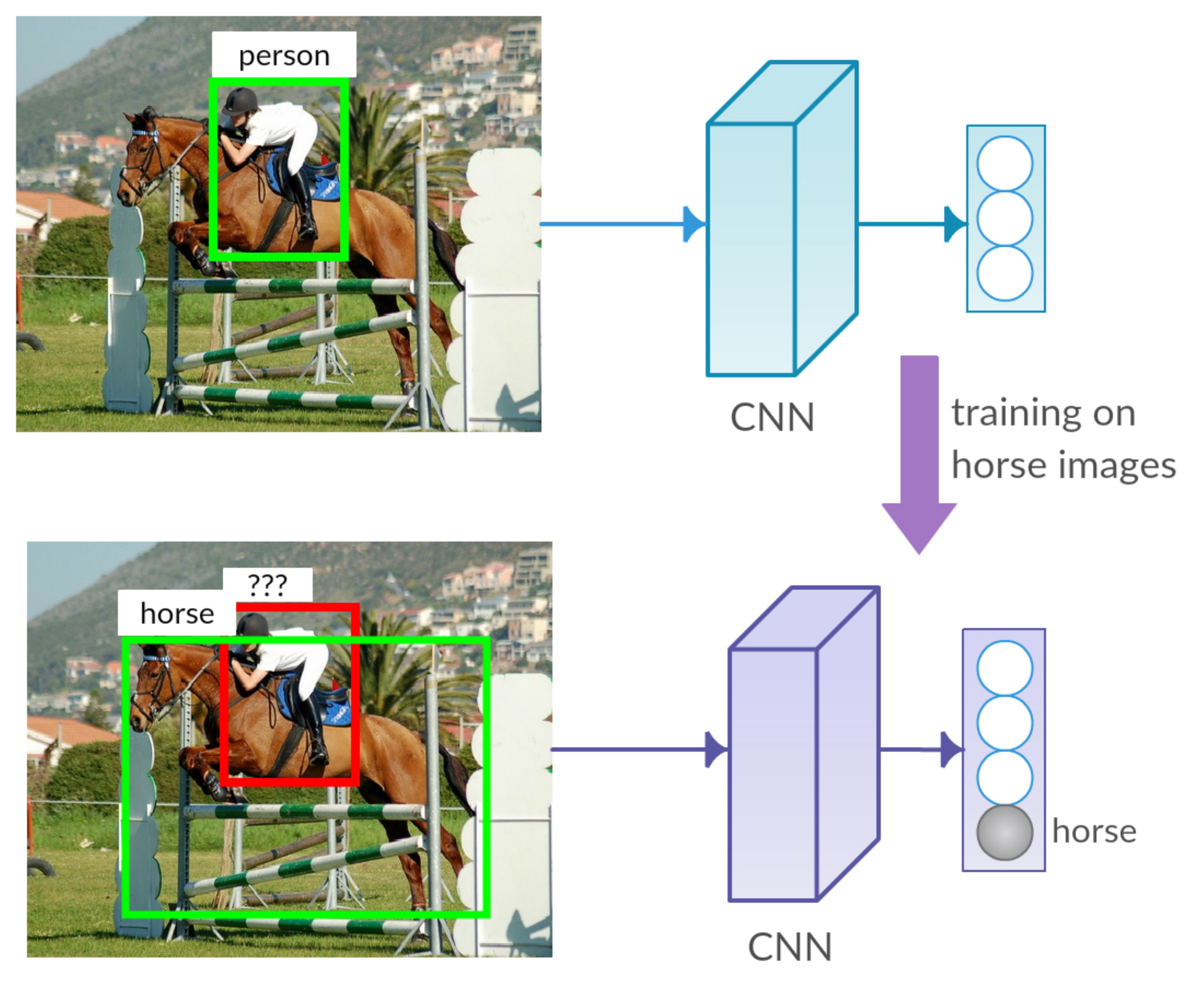}
\vspace{-0.2cm}
\caption{Catastrophic forgetting. An object detector network originally trained
for three classes, including {\it person}, detects the rider (top). When the
network is retrained with images of the new class {\it horse}, it detects the
horse in the test image, but fails to localize the rider (bottom).}
\vspace{-0.9cm}
\label{fig:teaser}
\end{center}
\end{figure}

\section{Introduction} \label{sec:intro}
\vspace{-0.2cm}
Modern detection methods, such as~\cite{ren2015faster,bell2015inside}, based on
convolutional neural networks (CNNs) have achieved state-of-the-art results on
benchmarks such as PASCAL VOC~\cite{pascal} and COCO~\cite{mscoco}. This,
however, comes with a high training time to learn the models. Furthermore, in
an era where datasets are evolving regularly, with new classes and samples, it is
necessary to develop incremental learning methods. A popular way to mitigate
this is to use CNNs pretrained on a certain dataset for a task, and adapt them
to new datasets or tasks, rather than train the entire network from scratch.

Fine-tuning~\cite{girshick2014rich} is one approach to adapt a network to new
data or tasks. Here, the output layer of the original network is adjusted,
either by replacing it with classes corresponding to the new task, or by adding
new classes to the existing ones. The weights in this layer are then randomly
initialized, and all the parameters of the network are tuned with the objective
for the new task. While this framework is very successful on the new classes,
its performance on the old ones suffers dramatically, if the network is not
trained on all the classes jointly. This issue, where a neural network forgets
previously learned knowledge when adapted to a new task, is referred to as
catastrophic interference or forgetting. It has been known for over a couple of
decades in the context of feedforward fully connected
networks~\cite{mccloskey1989catastrophic,ratcliff1990connectionist}, and needs
to be addressed in the current state-of-the-art object detector networks, if we
want to do incremental learning.

Consider the example in Figure~\ref{fig:teaser}. It illustrates catastrophic
forgetting when incrementally adding a class, {\it horse} in this object
detection example. The first CNN (top) is trained on three classes, including
{\it person}, and localizes the rider in the image. The second CNN (bottom) is
an incrementally trained version of the first one for the category {\it horse}.
In other words, the original network is adapted with images from only this new
class. This adapted network localizes the horse in the image, but fails to
detect the rider, which it was capable of originally, and despite the fact that
the {\it person} class was not updated.  In this paper, we present a method to
alleviate this issue.

\begin{figure*}[t]
\begin{center}
\includegraphics[width=0.85\linewidth]{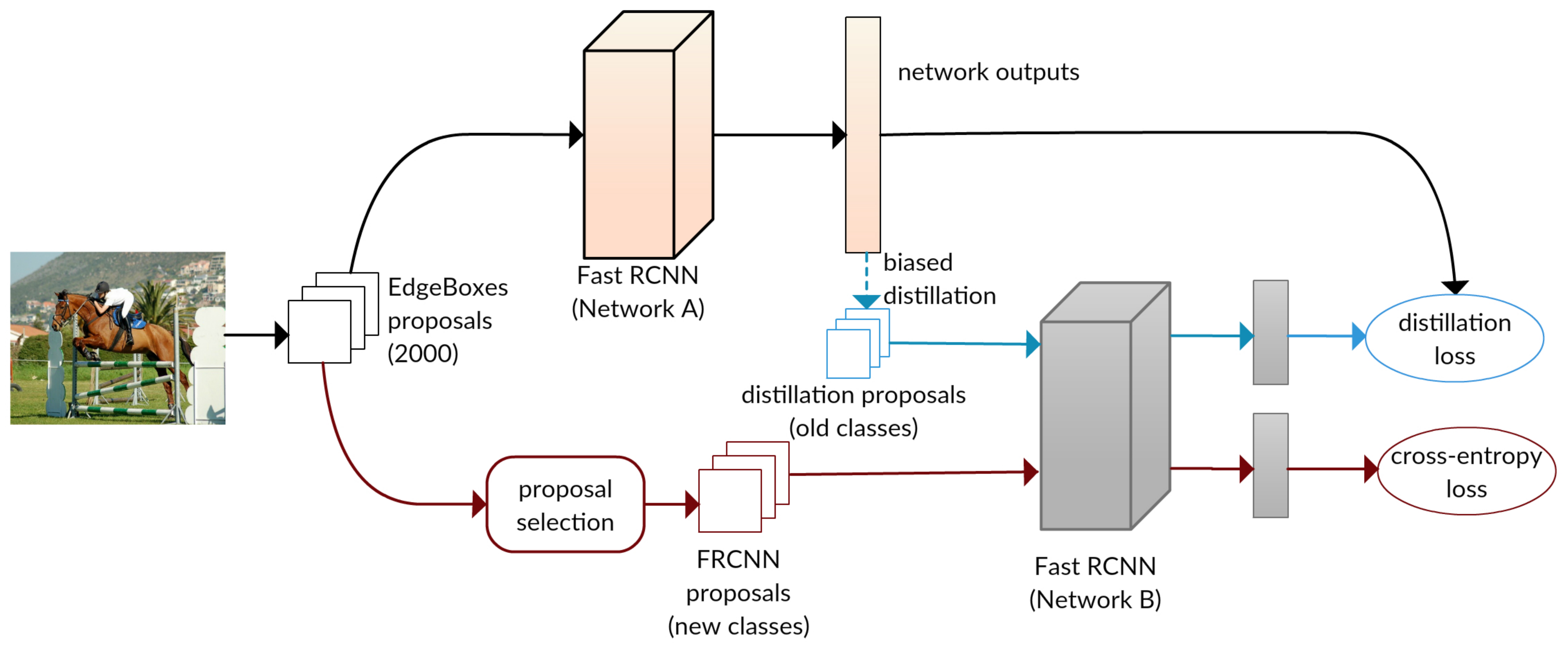}
\end{center}
\vspace{-0.5cm}
\caption{Overview of our framework for learning object detectors incrementally.
It is composed of a frozen copy of the detector (Network A) and the detector
(Network B) adapted for the new class(es). See text for details.}
\vspace{-0.5cm}
\label{fig:schema}
\end{figure*}

Using only the training samples for the new classes, we propose a method for
not only adapting the old network to the new classes, but also ensuring
performance on the old classes does not degrade. The core of our approach is a
loss function balancing the interplay between predictions on the new classes,
i.e., cross-entropy loss, and a new distillation loss which minimizes the
discrepancy between responses for old classes from the original and the new
networks. The overall approach is illustrated in Figure~\ref{fig:schema}.

We use a frozen copy of the original detection network to compute the
distillation loss. This loss is related to the concept of ``knowledge
distillation'' proposed in~\cite{hinton2015distilling}, but our application of
it is significantly different from this previous work, as discussed in
Section~\ref{sec:dual_net}. We specifically target the problem of object
detection, which has the additional challenge of localizing objects with
bounding boxes, unlike other attempts~\cite{li2016learning,icarl} limited to
the image classification task. We demonstrate experimental results on
the PASCAL VOC and COCO datasets using Fast R-CNN~\cite{girshick2015fast} as
the network. Our results show that we can add new classes incrementally to an
existing network without forgetting the original classes, and with no access to
the original training data. We also evaluate variants of our method
empirically, and show the influence of distillation and the loss function.
Note that our framework is general and can be applied to any other CNN-based
object detectors where proposals are computed externally, or static sliding
windows are used.

\vspace{-0.1cm}
\section{Related work} \label{sec:related}
\vspace{-0.1cm}

The problem of incremental learning has a long history in machine learning and
artificial
intelligence~\cite{thrun1996learning,Schlimmer86,polikar2001learn,Cauwenberghs00}.
Some of the more recent work, e.g.,~\cite{chen2013neil,levan}, focuses on
continuously updating the training set with data acquired from the Internet.
They are: (i) restricted to learning with a fixed data
representation~\cite{levan}, or (ii) keep all the collected data to retrain the
model~\cite{chen2013neil}. Other work partially addresses these issues by
learning classifiers without access to the ensemble of
data~\cite{Mensink13,Ristin14}, but uses a fixed image representation. Unlike
these methods, our approach is aimed at learning the representation and
classifiers jointly, without storing all the training examples. To this end, we
use neural networks to model the task in an end-to-end fashion.

Our work is also topically related to transfer learning and domain adaptation
methods. Transfer learning uses knowledge acquired from one task to help learn
another. Domain adaptation transfers the knowledge acquired for a task from a
data distribution to other (but related) data. These paradigms, and in
particular fine-tuning, a special case of transfer learning, are very popular
in computer vision. CNNs learned for image
classification~\cite{krizhevsky2012imagenet} are often used to train other
vision tasks such as object
detection~\cite{oquab2014learning,yosinski2014transferable} and semantic
segmentation~\cite{chen2014semantic}.

An alternative to transfer knowledge from one network to another is
distillation~\cite{hinton2015distilling,bucilu2006model}. This was originally
proposed to transfer knowledge between different neural networks---from a large
network to a smaller one for efficient deployment. The method
in~\cite{hinton2015distilling} encouraged the large (old) and the small (new)
networks to produce similar responses. It has found several applications in
domain adaptation and model
compression~\cite{gupta2015cross,rusu2015policy,tzeng2015simultaneous}.
Overall, transfer learning and domain adaptation methods require at least
unlabeled data for both the tasks or domains, and in its absence, the new
network quickly forgets all the knowledge acquired in the source
domain~\cite{french1999catastrophic,mccloskey1989catastrophic,ratcliff1990connectionist,goodfellow2013empirical}.
In contrast, our approach addresses the challenging case where no training data
is available for the original task (i.e., detecting objects belonging to the
original classes), by building on the concept of knowledge
distillation~\cite{hinton2015distilling}.

This phenomenon of forgetting is believed to be caused by two
factors~\cite{french1994dynamically,lecun1989backpropagation}. First, the
internal representations in hidden layers are often overlapping, and a small
change in a single neuron can affect multiple representations at the same
time~\cite{french1994dynamically}. Second, all the parameters in feedforward
networks are involved in computations for every data point, and a
backpropagation update affects all of them in each training
step~\cite{lecun1989backpropagation}. The problem of addressing these issues in
neural networks has its origin in classical connectionist networks several
years
ago~\cite{french1999catastrophic,mccloskey1989catastrophic,french1994dynamically,french2001pseudopatterns,ans2004self},
but needs to be adapted to today's large deep neural network architectures for
vision tasks~\cite{li2016learning,icarl}.

Li and Hoiem~\cite{li2016learning} use knowledge distillation for one of the
classical vision tasks, image classification, formulated in a deep learning
framework. However, their evaluation is limited to the case where the old
network is trained on a dataset, while the new network is trained on a
different one, e.g., Places365 for the old and PASCAL VOC for the new, ImageNet
for the old and PASCAL VOC for the new, etc.  While this is interesting, it is
a simpler task, because: (i) different datasets often contain dissimilar
classes, (ii) there is little confusion between datasets---it is in fact
possible to identify a dataset simply from an
image~\cite{torralba2011unbiased}.

Our method is significantly different from~\cite{li2016learning} in two ways.
First, we deal with the more difficult problem of learning incrementally on the
same dataset, i.e., the addition of classes to the network. As shown
in~\cite{icarl},~\cite{li2016learning} fails in a similar setting of learning
image classifiers incrementally. Second, we address the object detection task,
where it is very common for the old and the new classes to co-occur, unlike the
classification task.

Very recently, Rebuffi~\etal~\cite{icarl} address some of the drawbacks
in~\cite{li2016learning} with their incremental learning approach for image
classification. They also use knowledge distillation, but decouple the
classifier and the representation learning. Additionally, they rely on a subset
of the original training data to preserve the performance on the old classes.
In comparison, our approach is an end-to-end learning framework, where the
representation and the classifier are learned jointly, and we do not use any of
the original training samples to avoid catastrophic forgetting.
Alternatives to distillation are: growing the capacity of the network with new
layers~\cite{rusu2016progressive}, applying strong per-parameter regularization
selectively~\cite{kirkpatrick14032017}. The downside to these methods is the
rapid increase in the number of new parameters to be
learned~\cite{rusu2016progressive}, and their limited evaluation on the easier
task of image classification~\cite{kirkpatrick14032017}.

In summary, none of the previous work addresses the problem of learning
classifiers for object detection incrementally, without using previously seen
training samples.

\vspace{-0.2cm}
\section{Incremental learning of new classes} \label{sec:inclearn}
\vspace{-0.2cm}
Our overall approach for incremental learning of a CNN model for object
detection is illustrated in Figure~\ref{fig:schema}. It contains a frozen copy
of the original detector (denoted by Network A in the figure), which is used
to: (i) select proposals corresponding to the old classes, i.e., distillation
proposals, and (ii) compute the distillation loss. Network B in the figure is
the adapted network for the new classes. It is obtained by increasing the
number of outputs in the last layer of the original network, such that the new
output layer includes the old as well as the new classes.

In order to avoid catastrophic forgetting, we constrain the learning process of
the adapted network. We achieve this by incorporating a distillation loss, to
preserve the performance on the old classes, as an additional term in the
standard cross-entropy loss function (see \S\ref{sec:dual_net}). Specifically,
we evaluate each new training sample on the frozen copy (Network A) to choose a
diverse set of proposals (distillation proposals in Figure~\ref{fig:schema}),
and record their responses. With these responses in hand, we compute a
distillation loss which measures the discrepancy between the two networks for
the distillation proposals. This loss is added to the cross-entropy loss on the
new classes to make up the loss function for training the adapted detection
network. As we show in the experimental evaluation, the distillation loss as
well as the strategy to select the distillation proposals are critical in
preserving the performance on the old classes (see \S\ref{sec:expts}).

In the remainder of this section, we provide details of the object detector
network (\S\ref{sec:frcnn}), the loss functions and the learning algorithm
(\S\ref{sec:dual_net}), and strategies to sample the object proposals
(\S\ref{sec:sampling}).

\vspace{-0.1cm}
\subsection{Object detection network} \label{sec:frcnn}
\vspace{-0.2cm}
We use a variant of a popular framework for object detection---Fast
R-CNN~\cite{girshick2015fast}, which is a proposal-based detection method built
with pre-computed object proposals,
e.g.,~\cite{zitnick2014edge,arbelaez2014multiscale}. We chose this instead of
the more recent Faster R-CNN~\cite{ren2015faster}, which integrates the
computation of category-specific proposals into the network, because we need
proposals agnostic to object categories, such as
EdgeBoxes~\cite{zitnick2014edge}, MCG~\cite{arbelaez2014multiscale}. We use
EdgeBoxes~\cite{zitnick2014edge} proposals for PASCAL VOC 2007 and
MCG~\cite{arbelaez2014multiscale} for COCO. This allows us to focus on the
problem of learning the representation and the classifier, given a pre-computed
set of generic object proposals.

In our variant of Fast R-CNN, we replaced the VGG-16 trunk with a deeper
ResNet-50~\cite{he2015deep} component, which is faster and more accurate than
VGG-16. We follow the suggestions in~\cite{he2015deep} to combine Fast R-CNN
and ResNet architectures. The network processes the whole image through a
sequence of residual blocks. Before the last strided convolution layer we
insert a RoI pooling layer, which performs maxpooling over regions of varied
sizes, i.e., proposals, into a $7\times 7$ feature map. Then we add the
remaining residual blocks, a layer for average pooling over spatial dimensions,
and two fully connected layers: a softmax layer for classification (PASCAL or
COCO classes, for example, along with the background class) and a regression
layer for bounding box refinement, with independent corrections for each
class.

The input to the network is an image and about 2000 precomputed object
proposals represented as bounding boxes. During inference, the high-scoring
proposals are refined according to bounding box regression. Then, a
per-category non-maxima suppression (NMS) is performed to get the final
detection results. The loss function to train the Fast R-CNN detector,
corresponding to a RoI, is given by:
\vspace{-0.2cm}
\begin{equation}\label{eq:frcnn_loss}
    \mathcal{L}_{\text{rcnn}}(\mathbf{p}, k^*, t, t^*) = -\log \mathbf{p}_{k^*} + [k^* \geq 1] R(t - t^*),
\vspace{-0.2cm}
\end{equation}
where $\mathbf{p}$ is the set of responses of the network for all the classes
(i.e., softmax output), $k^*$ is a groundtruth class, $t$ is an output of
bounding box refinement layer, and $t*$ is the ground truth bounding box
proposal. The first part of the loss denotes log-loss over classes, and the
second part is localization loss. For more implementation details about Fast
R-CNN, refer to the original paper~\cite{girshick2015fast}.

\vspace{-0.1cm}
\subsection{Dual-network learning}\label{sec:dual_net}
\vspace{-0.1cm}
First, we train a Fast R-CNN to detect the original set of classes $C_A$. We
refer to this network as $\mathbf{A}(C_A)$. The goal now is to add a new set of
classes $C_B$ to this. We make two copies of $\mathbf{A}(C_A)$: one that is
frozen to recognize classes $C_A$ through distillation loss, and the second
$\mathbf{B}(C_B)$ that is extended to detect the new classes $C_B$, which were
not present or at least not annotated in the source images. The extension is
done only in the last fully connected layers, i.e., classification and bounding
box regression. We create sibling (i.e., fully-connected)
layers~\cite{girshick2014rich} for new classes only and concatenate their
outputs with the original ones. The new layers are initialized randomly in the
same way as the corresponding layers in Fast R-CNN.  Our goal is to train
$\mathbf{B}(C_B)$ to recognize classes $C_A \cup C_B$ using only new data and
annotations for $C_B$.

The distillation loss represents the idea of ``keeping all the answers of the
network the same or as close as possible''. If we train $\mathbf{B}(C_B)$
without distillation, average precision on the old classes will degrade
quickly, after a few hundred SGD iterations. This is a manifestation of
catastrophic forgetting. We illustrate this in Sections~\ref{sec:19p1}
and~\ref{sec:10p10}. We compute the distillation loss by applying the frozen
copy of $\mathbf{A}(C_A)$ to any new image. Even if no object is detected by
$\mathbf{A}(C_A)$, the unnormalized logits (softmax input) carry enough
information to ``distill'' the knowledge of the old classes from
$\mathbf{A}(C_A)$ to $\mathbf{B}(C_B)$. This process is illustrated in
Figure~\ref{fig:schema}.

For each image we randomly sample 64 RoIs out of 128 with the smallest
background score. The logits computed for these RoIs by $\mathbf{A}(C_A)$ serve
as targets for the old classes in the $L_2$ distillation loss shown below. The
logits for the new classes $C_B$ are not considered in this loss. We subtract
the mean over the class dimension from these unnormalized logits ($y$) of each
RoI to obtain the corresponding centered logits $\bar{y}$ used in the
distillation loss. Bounding box regression outputs $t_A$ (of the same set of
proposals used for computing the logit loss) also constrain the loss of the
network $\mathbf{B}(C_B)$. We chose to use $L_2$ loss instead of a
cross-entropy loss for regression outputs because it demonstrates more stable
training and performs better (see \S\ref{sec:10p10}). The distillation loss
combining the logits and regression outputs is written as:
\vspace{-0.2cm}
\begin{equation}\label{eq:distillation_loss}
\begin{split}
    \mathcal{L}_{\text{dist}}(y_A, t_A, y_B, t_B) &= \frac{1}{N|C_A|} \sum \Big[ (\bar{y}_A - \bar{y}_B)^2 \\ &+ (t_A - t_B)^2 \Big],
\vspace{-0.4cm}
\end{split}
\end{equation}
where $N$ is the number of RoIs sampled for distillation (i.e., 64 in this
case), $|C_A|$ is the number of old classes, and the sum is over all the RoIs
for the old classes. We distill logits without any smoothing,
unlike~\cite{hinton2015distilling},
because most of the proposals already produce a smooth distribution of scores.
Moreover, in our case, both the old and the new networks are similar with
almost the same parameters (in the beginning), and so smoothing the logits
distribution is not necessary to stabilize the learning.

The values of the bounding box regression are also distilled because we update
all the layers, and any update of the convolutional layers will affect them
indirectly. As box refinements are important to detect objects accurately,
their values should be conserved as well. This is an easier task than keeping
the classification scores because bounding box refinements for each class are
independent, and are not linked by the softmax.

The overall loss $\mathcal{L}$ to train the model incrementally is a weighted
sum of the distillation loss (\ref{eq:distillation_loss}), and the standard
Fast R-CNN loss (\ref{eq:frcnn_loss}) that is applied only to new classes $C_B$,
where groundtruth bounding box annotation is available. In essence,
\vspace{-0.1cm}
\begin{equation}\label{eq:joint_loss}
    \mathcal{L} = \mathcal{L}_{\text{rcnn}}+ \lambda \mathcal{L}_{\text{dist}},
\vspace{-0.1cm}
\end{equation}
where the hyperparameter $\lambda$ balances the two losses. We set $\lambda$ to
1 in all the experiments with cross-validation (see \S\ref{sec:10p10}).

The interplay between the two networks $\mathbf{A}(C_A)$ and $\mathbf{B}(C_B)$
provides the necessary supervision that prevents the catastrophic forgetting in
the absence of original training data used by $\mathbf{A}(C_A)$. After the
training of $\mathbf{B}(C_B)$ is completed, we can add more classes by freezing
the newly trained network and using it for distillation. We can thus add new
classes sequentially. Since $\mathbf{B}(C_B)$ is structurally identical to
$\mathbf{A}(C_A \cup C_B)$, the extension can be repeated to add more classes.

\vspace{-0.1cm}
\subsection{Sampling strategy}\label{sec:sampling}
\vspace{-0.1cm}
As mentioned before, we choose 64 proposals out of 128 with the lowest
background score, thus biasing the distillation to non-background proposals. We
noticed that proposals recognized as confident background do not provide strong
learning cues to conserve the original classes. One possibility is using an
\emph{unbiased distillation} that randomly samples 64 proposals out of the
whole set of 2000 proposals. However, when doing so, the detection performance
on old classes is noticeably worse because most of the distillation proposals
are now background, and carry no strong signal about the object categories.
Therefore, it is advantageous to select non-background proposals. We
demonstrate this empirically in Section~\ref{sec:inc_add}.

\vspace{-0.1cm}
\section{Experiments} \label{sec:expts}
\vspace{-0.1cm}

\subsection{Datasets and evaluation} \label{sec:dataset}
\vspace{-0.1cm}
We evaluate our method on the PASCAL VOC 2007 detection benchmark and the
Microsoft COCO challenge dataset. VOC 2007 consists of 5K images in the
trainval split and 5K images in the test split for 20 object classes. COCO on
the other hand has 80K images in the training set and 40K images in the
validation set for 80 object classes (which includes all the classes from VOC).
We use the standard mean average precision (mAP) at 0.5 IoU threshold as the
evaluation metric. We also report mAP weighted across different $\iou$ from
$0.5$ to $0.95$ on COCO, as recommended in the COCO challenge guidelines.
Evaluation of the VOC 2007 experiments is done on the test split, while for
COCO, we use the first 5000 images from the validation set.

\begin{table}[t!]
\centering
\begin{tabular}{l|c c c}
method & old & new & all\\
  \hline
$\mathbf{A}$(1-19) & 68.4 & - & - \\
+$\mathbf{B}$(20) w/o distillation & 25.0 & 52.1 & 26.4 \\
+$\mathbf{B}$(20) w frozen trunk & 53.5 & 43.1 & 52.9 \\
+$\mathbf{B}$(20) w all layers frozen & 69.1 & 41.6 & 66.6 \\
+$\mathbf{B}$(20) w frozen trunk and distill. & 68.7 & 43.2 & 67.4 \\
+$\mathbf{B}$(20) w distillation & 68.3 & 58.3 & 67.8 \\
+$\mathbf{B}$(20) w cross-entropy distill. & 68.1 & 52.0 & 67.3 \\
+$\mathbf{B}$(20) w/o bbox distillation & 68.5 & 62.7 & 68.3 \\
  \hline
$\mathbf{A}$(1-20) & 69.6 & 73.9 & 69.8 \\
\end{tabular}
\caption{\textbf{VOC 2007 test} average precision (\%). Experiments
demonstrating the addition of ``tvmonitor'' class to a pretrained network under
various setups. Classes 1-19 are the old classes, and ``tvmonitor'' (class 20)
is the new one.}
\vspace{-0.5cm}
\label{tab:voc19p1_compact}
\end{table}

\vspace{-0.1cm}
\subsection{Implementation details}
\label{sec:implement}
\vspace{-0.1cm}
We use SGD with Nesterov momentum~\cite{nesterov1983method} to train the
network in all the experiments. We set the learning rate to 0.001, decay to
0.0001 after 30K iterations, and momentum to 0.9. In the second stage of
training, i.e., learning the extended network with new classes, we used a
learning rate of 0.0001. The $\mathbf{A}(C_A)$ network is trained for 40K
iterations on PASCAL VOC 2007 and for 400K iterations on COCO. The
$\mathbf{B}(C_B)$ network is trained for 3K-5K iterations when only one class
is added, and for the same number of iterations as $\mathbf{A}(C_A)$
when many classes are added at once. Following Fast
R-CNN~\cite{girshick2015fast}, we regularize with weight decay of 0.00005 and
take batches of two images each. All the layers of $\mathbf{A}(C_A)$
and $\mathbf{B}(C_B)$ networks are finetuned unless stated otherwise.

The integration of ResNet into Fast R-CNN (see \S\ref{sec:frcnn}) is done
by adding a RoI pooling layer before the conv5\_1 layer, and replacing
the final classification layer by two sibling fully connected layers. The
batch normalization layers are frozen, and as in Fast R-CNN, no dropout is used.
RoIs are considered as detections if they have a score more than $0.5$ for any
of the classes. We apply per-class NMS with an $\iou$ threshold of $0.3$.
Training is image-centric, and a batch is composed of 64 proposals per image,
with 16 of them having an $\iou$ of at least $0.5$ with a groundtruth object.
All the proposals are filtered to have IoU less than $0.7$, as
in~\cite{zitnick2014edge}.

We use TensorFlow~\cite{tensorflow} to develop our incremental learning
framework. Each experiment begins with choosing a subset of classes to form the
set $C_A$. Then, a network is learned only on the subset of the training set
composed of all the images containing at least one object from $C_A$.
Annotations for other classes in these images are ignored. With the new classes
chosen to form the set $C_B$, we learn the extended network as described in
Section~\ref{sec:dual_net} with the subset of the training set containing at
least one object from $C_B$. As in the previous case, annotations of all the
other classes, including those of the original classes $C_A$, are ignored. For
computational efficiency, we precomputed the responses of the frozen network
$\mathbf{A}(C_A)$ on the training data (as every image is typically used
multiple times).

\vspace{-0.1cm}
\subsection{Addition of one class}\label{sec:19p1}
\vspace{-0.1cm}
In the first experiment we take 19 classes in alphabetical order from the VOC
dataset as $C_A$, and the remaining one as the only new class $C_B$. We then
train the $\mathbf{A}$(1-19) network on the VOC trainval subset containing any
of the 19 classes, and the $\mathbf{B}$(20) network is trained on the trainval
subset containing the new class. A summary of the evaluation of these networks
on the VOC test set is shown in Table~\ref{tab:voc19p1_compact}, with the full
results in Table~\ref{tab:19p1_voc_exp}.

A baseline approach for addition of a new class is to add an output to the last
layer and freeze the rest of the network. This freezing, where the weights of
the network's convolutional layers are fixed (``$\mathbf{B}$(20) w frozen
trunk'' in the tables), results in a lower performance on the new class as the
previously learned representations have not been adapted for it. Furthermore,
it does not prevent degradation of the performance on the old classes, where
mAP drops by almost 15\%. When we freeze all the layers, including the old
output layer (``$\mathbf{B}$(20) w all layers frozen''), or apply distillation
loss (``$\mathbf{B}$(20) w frozen trunk and distill.''), the performance on the
old classes is maintained, but that on the new class is poor. This shows that
finetuning of convolutional layers is necessary to learn the new classes.

\begin{table}[t!]
\centering
\begin{tabular}{l|c c c}
method & old & new & all\\
  \hline
$\mathbf{A}$(1-10) & 65.8 & - & -\\
+$\mathbf{B}$(11-20) w/o distillation & 12.8 & 64.5 & 38.7 \\
+$\mathbf{B}$(11-20) w distillation & 63.2 & 63.1 & 63.1\\
+$\mathbf{B}$(11-20) w/o bbox distillation & 58.7 & 63.1 & 60.9 \\
+$\mathbf{B}$(11-20) w EWC~\cite{kirkpatrick14032017} & 31.6 & 61.0 & 46.3\\
\hline
$\mathbf{A}$(1-20) & 68.4 & 71.3 & 69.8 \\
\end{tabular}
\caption{\textbf{VOC 2007 test} average precision (\%). Experiments
demonstrating the addition of 10 classes, all at once, to a pretrained network.
Classes 1-10 are the old classes, and 11-20 the new ones.}
\vspace{-0.4cm}
\label{tab:voc10p10_compact}
\end{table}

When the network $\mathbf{B}$(20) is trained without the distillation loss
(``$\mathbf{B}$(20) w/o distillation'' in the tables), it can learn the 20th
class, but the performance decreases significantly on the other (old) classes.
As seen in Table~\ref{tab:19p1_voc_exp}, the AP on classes like ``cat'',
``person'' drops by over 60\%. The same training procedure with distillation
loss largely alleviates this catastrophic forgetting. Without distillation, the
new network has 25.0\% mAP on the old classes compared to 68.3\% with
distillation, and 69.6\% mAP of baseline Fast R-CNN trained jointly on all
classes (``A(1-20)''). With distillation the performance is similar to that of
the old network $\mathbf{A}$(1-19), but is lower for certain classes, e.g.,
``bottle''. The 20th class ``tvmonitor'' does not get the full performance of
the baseline (73.9\%), with or without distillation, and is less than 60\%.
This is potentially due to the size of the training set. The $\mathbf{B}(20)$
network is trained only a few hundred images containing instances of this
class. Thus, the ``tvmonitor'' classifier does not see the full diversity of
negatives.

We also performed the ``addition of one class'' experiment with each of
the VOC categories being the new class. The behavior for each class is very
similar to the ``tvmonitor'' case described above. The mAP varies from 66.1\%
(for new class ``sheep'') to 68.3\% (``tvmonitor'') with mean 67.38\% and
standard deviation of 0.6\%.

\vspace{-0.1cm}
\subsection{Addition of multiple classes}\label{sec:10p10}
\vspace{-0.1cm}
In this scenario we train the network $\mathbf{A}$(1-10) on the first 10 VOC
classes (in alphabetical order) with the VOC trainval subset corresponding to
these classes. In the second stage of training we used the remaining 10 classes
as $C_B$ and trained only on the images containing the new classes.
Table~\ref{tab:voc10p10_compact} shows a summary of the evaluation of these
networks on the VOC test set, with the full results in
Table~\ref{tab:10p10_voc_exp}.

Training the network $\mathbf{B}$(11-20) on the 10 new classes with
distillation (for the old classes) achieves 63.1\% mAP (``$\mathbf{B}$(11-20) w
distillation'' in the tables) compared to 69.8\% of the baseline network
trained on all the 20 classes (``$\mathbf{A}$(1-20)''). Just as in the previous
experiment of adding one class, performance on the new classes is slightly
worse than with the joint training of all the classes. For example, as seen in
Table~\ref{tab:10p10_voc_exp}, the performance for ``person'' is 73.2\% vs
79.1\%, and 72.5\% vs 76.8\% for the ``train'' class. The mAP on new classes is
63.1\% for the network with distillation versus 71.3\% for the jointly trained
model. However, without distillation, the network achieves only 12.8\% mAP
(``+$\mathbf{B}$(11-20) w/o distillation'') on the old classes. Note that the
method without bounding box distillation (``+$\mathbf{B}$(11-20) w/o bbox
distillation'') is inferior to our full method (``+$\mathbf{B}$(11-20) w
distillation'').

\begin{figure}[t]
\begin{center}
  \includegraphics[width=0.75\linewidth]{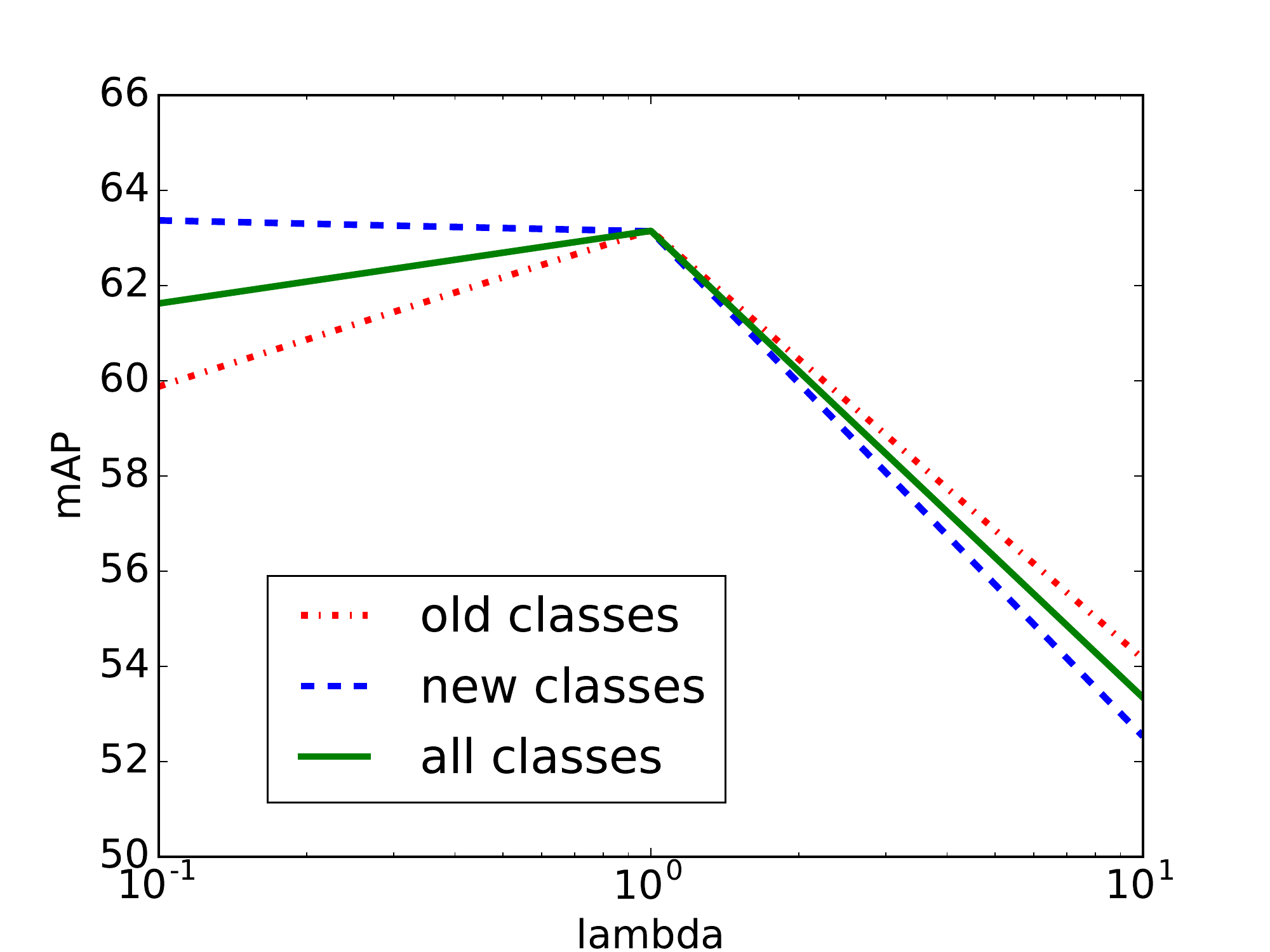}
\end{center}
\vspace{-0.4cm}
\caption{The influence of $\lambda$, in the loss function
(\ref{eq:joint_loss}), on the mAP performance for the $\mathbf{B}$(11-20)
network trained with distillation.}
\vspace{-0.5cm}
\label{fig:lambda}
\end{figure}

We also performed the 10-class experiment for different values of $\lambda$ in
(\ref{eq:joint_loss}), the hyperparameter controlling the relative importance
of distillation and Fast R-CNN loss. Results shown in Figure~\ref{fig:lambda}
demonstrate that when the distillation is weak ($\lambda=0.1$) the new classes
are easier to learn, but the old ones are more easily forgotten. When
distillation is strong ($\lambda=10$), it destabilizes training and impedes
learning the new classes. Setting $\lambda$ to $1$ is a good trade-off between
learning new classes and preventing catastrophic forgetting.

\begin{table}[t!]
\centering
\begin{tabular}{l|c c c}
method & old & new & all\\
\hline
$\mathbf{A}$(1-15) & 70.5 & - & -\\
+$\mathbf{B}$(16-20) w distill. & 68.4 & 58.4 & 65.9\\
+$\mathbf{B}$(16)(17)...(20) w distill. & 66.0 & 51.6 & 62.4 \\
+$\mathbf{B}$(16)(17)...(20) w unbiased distill. & 45.8 & 46.5 & 46.0 \\
+$\mathbf{A}$(16)+...+$\mathbf{A}$(20) & 70.5 & 37.8 & 62.4\\
\hline
$\mathbf{A}$(1-20) & 70.9 & 66.7 & 69.8 \\
\end{tabular}
\caption{\textbf{VOC 2007 test} average precision (\%). Experiments
demonstrating the addition of 5 classes, all at once, and incrementally to a
pretrained network. Classes 1-15 are the old ones, and 16-20 the new classes.}
\vspace{-0.2cm}
\label{tab:voc15p5_compact}
\end{table}

\begin{table}[t!]
\renewcommand{\arraystretch}{1.2}
\renewcommand{\tabcolsep}{1.2mm}
\centering
\begin{tabular}{l|c|c}
method & mAP@.5 & mAP@[.5, .95] \\
\hline
$\mathbf{A}$(1-40)+$\mathbf{B}$(41-80) & 37.4 & 21.3\\
$\mathbf{A}$(1-80) & 38.1 & 22.6\\
\end{tabular}
\caption{\textbf{COCO minival} (first 5000 validation images) average precision
(\%). We compare the model learned incrementally on half the classes with the
baseline trained on all jointly.}
\vspace{-0.4cm}
\label{tab:coco_exp}
\end{table}

We also compare our approach with elastic weight consolidation
(EWC)~\cite{kirkpatrick14032017}, which is an alternative to distillation and
applies per-parameter regularization selectively to alleviate catastrophic
forgetting. We reimplemented EWC and verified that it produces results
comparable to those reported in~\cite{kirkpatrick14032017} on MNIST, and then
adapted it to our object detection task. We do this by using the Fast R-CNN
batches during the training phase (as done in Section~\ref{sec:implement}), and
by replacing log loss with the Fast R-CNN loss. Our approach outperforms EWC for
this case, when we add 10 classes at once, as shown in
Tables~\ref{tab:voc10p10_compact} and~\ref{tab:10p10_voc_exp}.

We evaluated the influence of the number of new classes in incremental
learning. To this end, we learn a network for 15 classes first, and then train
for the remaining 5 classes, all added at once on VOC. These results are
summarized in Table~\ref{tab:voc15p5_compact}, with the per-class results shown
in Table~\ref{tab:15p5_voc_exp}. The network $\mathbf{B}$(16-20) has better
overall performance than $\mathbf{B}$(11-20): 65.9\% mAP versus 63.1\% mAP.  As
in the experiment with 10 classes, the performance is lower for a few classes,
e.g., ``table'', ``horse'', for example, than the initial model
$\mathbf{A}$(1-15). The performance on the new classes is lower than jointly
trained baseline Fast R-CNN $\mathbf{A}$(1-20).
Overall, mAP of $\mathbf{B}$(16-20) is lower than baseline Fast R-CNN (65.9\%
versus 69.8\%).

The evaluation on COCO, shown in Table~\ref{tab:coco_exp}, is done with the
first 40 classes in the initial set, and the remaining 40 in the new second
stage. The network $\mathbf{B}$(41-80) trained with the distillation loss
obtains 37.4\% mAP in the PASCAL-style metric and 21\% mAP in the COCO-style
metric. The baseline network trained on 80 classes is similar in performance
with 38.1\% and 22.6\% mAP respectively. We observe that our proposed method
overcomes catastrophic forgetting, just as in the case of VOC seen earlier.

We also studied if distillation depends on the distribution of images used in
this loss. To this end, we used the model $\mathbf{A}$(1-10) trained on VOC,
and then performed the second stage learning in two settings:
$\mathbf{B}$(11-20) learned on the subset of VOC as before, and another model
trained for the same set of classes, but using a subset of COCO. From
Table~\ref{tab:voc10_coco10} we see that indeed, distillation works better when
background samples have exactly the same distribution in both stages of
training. However, it is still very effective even when the dataset in the
second stage is different from the one used in the first.

\vspace{-0.1cm}
\subsection{Sequential addition of multiple classes}\label{sec:inc_add}
\vspace{-0.1cm}
In order to evaluate incremental learning of classes added sequentially, we
update the frozen copy of the network with the one learned with the new class,
and then repeat the process with another new class. For example, we take a
network learned for 15 classes of VOC, train it for the 16th on the subset
containing only this class, and then use the 16-class network as the frozen
copy to then learn the 17th class. This is then continued until the 20th class.
We denote this incremental extension as $\mathbf{B}$(16)(17)(18)(19)(20).

Results of adding classes sequentially are shown in
Tables~\ref{tab:15p5_voc_exp} and~\ref{tab:15p5_voc_exp_intermediate}. After
adding the 5 classes we obtain 62.4\% mAP (row 3 in
Table~\ref{tab:15p5_voc_exp}), which is lower than 65.9\% obtained by adding
all the 5 classes at once (row 2). Table~\ref{tab:15p5_voc_exp_intermediate}
shows intermediate evaluations after adding each class. We observe that the
performance of the original classes remains stable at each step in most cases,
but for a few classes, which is not recovered in the following steps.
We empirically evaluate the importance of using biased non-background proposals
(cf.\ \S\ref{sec:sampling}). Here we add the 5 classes one by one, but use
unbiased distillation (``$\mathbf{B}$(16)(17)(18)(19)(20) w unbiased distill.''
in Tables~\ref{tab:voc15p5_compact} and~\ref{tab:15p5_voc_exp}), i.e., randomly
sampled proposals are used for distillation. This results in much worse overall
performance (46\% vs 62.4\%) and some classes (``person'', ``chair'') suffer
from a significant performance drop of 10-20\%.
We also performed sequential addition experiment with 10 classes, and present
the results in Table~\ref{tab:10p10_voc_exp_incremental}. Although the drop in
mAP is more significant than for the previous experiment with 5 classes, it is
far from catastrophic forgetting.

\begin{table}[t]
\centering
\begin{tabular}{l|r|r}
    & +COCO-10cls & +VOC-10cls\\
  \hline
  mAP (old classes) & 61.4 & 63.2\\
  mAP (new classes) & 48.2 & 63.1\\
  mAP (all classes) & 54.8 & 63.2\\
\end{tabular}
\caption{\textbf{VOC 2007 test} average precision (\%). The second stage of
training, where 10 classes (11-20th) are added, is done on the subset of COCO
images (+COCO-10cls), and is compared to the one trained on the VOC subset
(+VOC-10cls).}
\vspace{-0.5cm}
\label{tab:voc10_coco10}
\end{table}

\begin{table*}[t!]
\centering
\renewcommand{\arraystretch}{1.2}
\renewcommand{\tabcolsep}{1.2mm}
\resizebox{\linewidth}{!}{
\begin{tabular}{l|r|r|r|r|r|r|r|r|r|r|r|r|r|r|r|r|r|r|r|r|l}
method & aero & bike & bird & boat & bottle & bus & car & cat & chair & cow & table & dog & horse & mbike & persn & plant & sheep & sofa & train & tv & mAP \\
  \hline
$\mathbf{A}$(1-19) & 69.9 & 79.4 & 69.5 & 55.7 & 45.6 & 78.4 & 78.9 & 79.8 & 44.8 & 76.2 & 63.8 & 78.0 & 80.8 & 77.6 & 70.2 & 40.9 & 67.8 & 64.5 & 77.5 & - & 68.4 \\
+$\mathbf{B}$(20) w/o distillation & 35.9 & 36.1 & 26.4 & 16.5 & 9.1 & 26.4 & 36.2 & 18.2 & 9.1 & 51.5 & 9.1 & 26.6 & 50.0 & 26.2 & 9.1 & 9.1 & 43.7 & 9.1 & 28.0 & 52.1 & 26.4 \\
+$\mathbf{B}$(20) w frozen trunk & 61.3 & 71.9 & 62.5 & 46.2 & 34.5 & 70.6 & 71.6 & 62.4 & 9.1 & 68.3 & 27.1 & 61.6 & 80.0 & 70.6 & 35.9 & 24.6 & 53.8 & 34.9 & 68.9 & 43.1 & 52.9\\
  +$\mathbf{B}$(20) w all layers frozen & 68.8 & 78.4 & 70.2 & 51.8 & 52.8 & 76.1 & 78.7 & 78.8 & 50.1 & 74.5 & 65.5 & 76.9 & 80.2 & 76.3 & 69.8 & 40.4 & 62.0 & 63.7 & 75.5 & 41.6 & 66.6\\
+$\mathbf{B}$(20) w frozen trunk and distill. & 74.4 & 78.1 & 69.8 & 54.7 & 52.1 & 75.7 & 79.0 & 78.5 & 48.5 & 74.4 & 62.3 & 77.0 & 80.2 & 77.2 & 69.7 & 44.5 & 68.6 & 64.5 & 74.7 & 43.2 & 67.4 \\
+$\mathbf{B}$(20) w distillation & 70.2 & 79.3 & 69.6 & 56.4 & 40.7 & 78.5 & 78.8 & 80.5 & 45.0 & 75.7 & 64.1 & 77.8 & 80.8 & 78.0 & 70.4 & 42.3 & 67.6 & 64.6 & 77.5 & 58.3 & 67.8 \\
+$\mathbf{B}$(20) w cross-entropy distill. & 69.1 & 79.1 & 69.5 & 52.8 & 45.4 & 78.1 & 78.9 & 79.5 & 44.8 & 75.5 & 64.2 & 77.2 & 80.8 & 77.9 & 70.2 & 42.7 & 66.8 & 64.6 & 76.1 & 52.0 & 67.3 \\
+$\mathbf{B}$(20) w/o bbox distillation & 69.4 & 79.3 & 69.5 & 57.4 & 45.4 & 78.4 & 79.1 & 80.5 & 45.7 & 76.3 & 64.8 & 77.2 & 80.8 & 77.5 & 70.1 & 42.3 & 67.5 & 64.4 & 76.7 & 62.7 & 68.3 \\
\hline
$\mathbf{A}$(1-20) & 70.2 & 77.9 & 70.4 & 54.1 & 47.4 & 78.9 & 78.6 & 79.8 & 50.8 & 75.9 & 65.6 & 78.0 & 80.5 & 79.1 & 76.3 & 47.7 & 69.3 & 65.6 & 76.8 & 73.9 & 69.8 \\
\end{tabular}
}
\caption{\textbf{VOC 2007 test} per-class average precision (\%) under
different settings when the ``tvmonitor'' class is added.}
\vspace{-0.1cm}
\label{tab:19p1_voc_exp}
\end{table*}

\begin{table*}[t!]
\centering
\renewcommand{\arraystretch}{1.2}
\renewcommand{\tabcolsep}{1.2mm}
\resizebox{\linewidth}{!}{
\begin{tabular}{l|r|r|r|r|r|r|r|r|r|r|r|r|r|r|r|r|r|r|r|r|l}
method & aero & bike & bird & boat & bottle & bus & car & cat & chair & cow & table & dog & horse & mbike & persn & plant & sheep & sofa & train & tv & mAP \\
\hline
$\mathbf{A}$(1-10) & 69.9 & 76.7 & 68.9 & 54.9 & 48.7 & 72.9 & 78.8 & 75.5 & 48.8 & 62.7 & - & - & - & - & - & - & - & - & - & - & 65.8 \\
+$\mathbf{B}$(11-20) w/o distillation & 25.5 & 9.1 & 23.5 & 17.3 & 9.1 & 9.1 & 9.1 & 16.2 & 0.0 & 9.1 & 61.5 & 67.7 & 76.0 & 72.2 & 68.9 & 34.8 & 63.6 & 62.7 & 72.5 & 65.2 & 38.7\\
+$\mathbf{B}$(11-20) w distillation & 69.9 & 70.4 & 69.4 & 54.3 & 48.0 & 68.7 & 78.9 & 68.4 & 45.5 & 58.1 & 59.7 & 72.7 & 73.5 & 73.2 & 66.3 & 29.5 & 63.4 & 61.6 & 69.3 & 62.2 & 63.1\\
+$\mathbf{B}$(11-20) w/o bbox distillation & 68.8 & 69.8 & 60.6 & 46.4 & 46.7 & 65.9 & 71.3 & 66.3 & 43.6 & 47.3 & 58.5 & 70.6 & 73.4 & 70.6 & 66.3 & 33.6 & 63.1 & 62.1 & 69.4 & 63.1 & 60.9\\
+$\mathbf{B}$(11-20) w EWC~\cite{kirkpatrick14032017} & 54.5 & 18.2 & 52.8 & 20.8 & 25.8 & 53.2 & 45.0 & 27.3 & 9.1 & 9.1 & 49.6 & 61.2 & 76.1 & 73.6 & 67.1 & 35.8 & 57.8 & 55.2 & 67.9 & 65.3 & 46.3\\
\hline
$\mathbf{A}$(1-20) & 70.2 & 77.9 & 70.4 & 54.1 & 47.4 & 78.9 & 78.6 & 79.8 & 50.8 & 75.9 & 65.6 & 78.0 & 80.5 & 79.1 & 76.3 & 47.7 & 69.3 & 65.6 & 76.8 & 73.9 & 69.8 \\
\end{tabular}
}
\caption{\textbf{VOC 2007 test} per-class average precision (\%) under
different settings when 10 classes are added at once.}
\vspace{-0.1cm}
\label{tab:10p10_voc_exp}
\end{table*}

\begin{table*}[t!]
\centering
\renewcommand{\arraystretch}{1.2}
\renewcommand{\tabcolsep}{1.2mm}
\resizebox{\linewidth}{!}{
\begin{tabular}{l|r|r|r|r|r|r|r|r|r|r|r|r|r|r|r|r|r|r|r|r|l}
method & aero & bike & bird & boat & bottle & bus & car & cat & chair & cow & table & dog & horse & mbike & persn & plant & sheep & sofa & train & tv & mAP \\
  \hline
$\mathbf{A}$(1-15) & 70.8 & 79.1 & 69.8 & 59.2 & 53.3 & 76.9 & 79.3 & 79.1 & 47.8 & 70.0 & 62.0 & 76.6 & 80.4 & 77.5 & 76.2 & - &  - & - & - & - & 70.5\\
+$\mathbf{B}$(16-20) w distill. & 70.5 & 79.2 & 68.8 & 59.1 & 53.2 & 75.4 & 79.4 & 78.8 & 46.6 & 59.4 & 59.0 & 75.8 & 71.8 & 78.6 & 69.6 & 33.7 & 61.5 & 63.1 & 71.7 & 62.2 & 65.9\\
+$\mathbf{B}$(16)(17)(18)(19)(20) w distill. & 70.0 & 78.1 & 61.0 & 50.9 & 46.3 & 76.0 & 78.8 & 77.2 & 46.1 & 66.6 & 58.9 & 67.7 & 71.6 & 71.4 & 69.6 & 25.6 & 57.1 & 46.5 & 70.7 & 58.2 & 62.4\\
+$\mathbf{B}$(16)(17)(18)(19)(20) w unbiased distill. & 62.2 & 71.2 & 52.3 & 43.8 & 24.9 & 60.7 & 62.9 & 53.4 & 9.1 & 34.9 & 42.5 & 34.8 & 54.3 & 70.9 & 9.1 & 18.7 & 53.2 & 48.9 & 58.2 & 53.5 & 46.0\\
\hline
$\mathbf{A}$(1-20) & 70.2 & 77.9 & 70.4 & 54.1 & 47.4 & 78.9 & 78.6 & 79.8 & 50.8 & 75.9 & 65.6 & 78.0 & 80.5 & 79.1 & 76.3 & 47.7 & 69.3 & 65.6 & 76.8 & 73.9 & 69.8 \\
\end{tabular}
}
\caption{\textbf{VOC 2007 test} per-class average precision (\%) under
different settings when 5 classes are added at once or sequentially.}
\vspace{-0.1cm}
\label{tab:15p5_voc_exp}
\end{table*}

\begin{table*}[t!]
\centering
\renewcommand{\arraystretch}{1.2}
\renewcommand{\tabcolsep}{1.2mm}
\resizebox{\linewidth}{!}{
\begin{tabular}{l|r|r|r|r|r|r|r|r|r|r|r|r|r|r|r|r|r|r|r|r|l}
method & aero & bike & bird & boat & bottle & bus & car & cat & chair & cow & table & dog & horse & mbike & persn & plant & sheep & sofa & train & tv & mAP \\
  \hline
$\mathbf{A}$(1-15) & 70.8 & 79.1 & 69.8 & 59.2 & 53.3 & 76.9 & 79.3 & 79.1 & 47.8 & 70.0 & 62.0 & 76.6 & 80.4 & 77.5 & 76.2 & - &  - & - & - & - & 70.5\\
+$\mathbf{B}$(16) & 70.5 & 78.3 & 69.6 & 60.4 & 52.4 & 76.8 & 79.4 & 79.2 & 47.1 & 70.2 & 56.7 & 77.0 & 80.3 & 78.1 & 70.0 & 26.3 & - & - & - & - & 67.0\\
+$\mathbf{B}$(16)(17) & 70.3 & 78.9 & 67.7 & 59.2 & 47.0 & 76.3 & 79.3 & 77.7 & 48.0 & 58.8 & 60.2 & 67.4 & 71.6 & 78.6 & 70.2 & 27.9 & 46.8 & - & - & - & 63.9\\
+$\mathbf{B}$(16)(17)(18) & 69.8 & 78.2 & 67.0 & 50.4 & 46.9 & 76.5 & 78.6 & 78.0 & 46.4 & 58.6 & 58.6 & 67.5 & 71.8 & 78.5 & 69.9 & 26.1 & 56.2 & 45.3 & - & - & 62.5\\
+$\mathbf{B}$(16)(17)(18)(19) & 70.4 & 78.8 & 67.3 & 49.8 & 46.4 & 75.6 & 78.4 & 78.0 & 46.0 & 59.5 & 59.2 & 67.2 & 71.8 & 71.3 & 69.8 & 25.9 & 56.1 & 48.2 & 65.0 & - & 62.4\\
+$\mathbf{B}$(16)(17)(18)(19)(20) & 70.0 & 78.1 & 61.0 & 50.9 & 46.3 & 76.0 & 78.8 & 77.2 & 46.1 & 66.6 & 58.9 & 67.7 & 71.6 & 71.4 & 69.6 & 25.6 & 57.1 & 46.5 & 70.7 & 58.2 & 62.4\\
\hline
$\mathbf{A}$(1-20) & 70.2 & 77.9 & 70.4 & 54.1 & 47.4 & 78.9 & 78.6 & 79.8 & 50.8 & 75.9 & 65.6 & 78.0 & 80.5 & 79.1 & 76.3 & 47.7 & 69.3 & 65.6 & 76.8 & 73.9 & 69.8 \\
\end{tabular}
}
\caption{\textbf{VOC 2007 test} per-class average precision (\%) when 5 classes
are added sequentially.}
\vspace{-0.1cm}
\label{tab:15p5_voc_exp_intermediate}
\end{table*}

\begin{table*}[t!]
\centering
\renewcommand{\arraystretch}{1.2}
\renewcommand{\tabcolsep}{1.2mm}
\begin{tabular}{l|c|c|c|c|c|c|c|c|c|c|c}
method & $\mathbf{A}$(1-10) & +table & +dog & +horse & +mbike & +persn & +plant & +sheep & +sofa & +train & +tv \\
  \hline
mAP  & 67.1 & 65.1 & 62.5 & 59.9 & 59.8 & 59.2 & 57.3 & 49.1 & 49.8 & 48.7 & 49.0\\
\end{tabular}
\caption{\textbf{VOC 2007 test} average precision (\%)
when adding 10 classes sequentially. Unlike other tables each column
here shows the mAP of a network trained on all the previous classes and
the new class. For example, the mAP shown for ``+dog'' is the result of
the network trained on the first ten classes, ``table'', and the new
class ``dog''.}
\vspace{-0.4cm}
\label{tab:10p10_voc_exp_incremental}
\end{table*}

\vspace{-0.1cm}
\subsection{Other alternatives}
\vspace{-0.1cm}
\paragraph{Learning multiple networks.} Another solution for learning
multiple classes is to train a new network for each class, and then combine
their detections. This is an expensive strategy at test time, as each network
has to be run independently, including the extraction of features. This may
seem like a reasonable thing to do as evaluation of object detection is done
independently for each class, However, learning is usually not independent.
Although we can learn a decent detection network for 10 classes, it is much
more difficult when learning single classes independently. To demonstrate this,
we trained a network for 1-15 classes and then separate networks for each of
the 16-20 classes. This results in 6 networks in total (row
``+$\mathbf{A}$(16)+...+$\mathbf{A}$(20)'' in Table~\ref{tab:voc15p5_compact}),
compared to incremental learning of 5 classes implemented with a single network
(``+$\mathbf{B}$(16)(17)...(20) w distill.''). The results confirm that new
classes are difficult to learn in isolation.

\vspace{-0.3cm}
\paragraph{Varying distillation loss.} As noted in~\cite{hinton2015distilling},
knowledge distillation can also be expressed as a cross-entropy loss. We
compared this with $L_2$-based loss on the one class extension experiment
(``$\mathbf{B}(20)$ w cross-entropy distill.'' in
Tables~\ref{tab:voc19p1_compact} and~\ref{tab:19p1_voc_exp}). Cross-entropy
distillation works as well as $L_2$ distillation keeping old classes intact
(67.3\% vs 67.8\%), but performs worse than $L_2$ on the new class
``tvmonitor'' (52\% vs 58.3\%). We also observed that cross-entropy is more
sensitive to the training schedule. According to~\cite{hinton2015distilling},
both formulations should be equivalent in the limit of a high smoothing factor
for logits (cf.\ \S\ref{sec:dual_net}), but our choice of not smoothing leads
to this different behavior.

\vspace{-0.2cm}
\paragraph{Bounding box regression distillation.} Addition of 10 classes
(Table~\ref{tab:voc10p10_compact}) without distilling bounding box regression
values performs consistently worse than the full distillation loss. Overall
$\mathbf{B}$(11-20) without distilling bounding box regression gets 60.9\% vs
63.1\% with the full distillation. However, on a few new classes the
performance can be higher than with the full distillation
(Table~\ref{tab:10p10_voc_exp}). This is also the case for $\mathbf{B}$(20)
without bounding box distillation (Table~\ref{tab:19p1_voc_exp}) that has
better performance on ``tvmonitor'' (62.7\% vs 58.3\%). This is not the
case when other categories are chosen as the new class. Indeed, bounding box
distillation shows an improvement of 2\% for the ``sheep'' class.

\vspace{-0.2cm}
\section{Conclusion}
\vspace{-0.2cm}
In this paper, we have presented an approach for incremental learning of object
detectors for new classes, without access to the training data corresponding to
the old classes. We address the problem of catastrophic forgetting in this
context, with a loss function that optimizes the performance on the new
classes, in addition to preserving the performance on the old classes. Our
extensive experimental analysis demonstrates that our approach performs well,
even in the extreme case of adding new classes one by one.
Part of future work is adapting our method to learned proposals, e.g.,
from RPN for Faster R-CNN~\cite{ren2015faster}, by reformulating RPN as a single
class detector that works on sliding window proposals. This requires adding
another term for RPN-based knowledge distillation in the loss function.

\vspace{0.2cm}
{\noindent {\bf Acknowledgments.}
This work was supported in part by the ERC advanced grant ALLEGRO, a Google
research award, and gifts from Facebook and Intel. We gratefully acknowledge
NVIDIA's support with the donation of GPUs used for this work.}

{\small
\bibliographystyle{ieee}
\bibliography{egbib}
}

\end{document}